\documentclass{SPAICE}
\usepackage{tikz}
\usepackage{quantikz}
\def\authorEmail{abhishek.singh@eurac.edu}
\def\AuthorShort{A. Singh et al.}
\author[1]{Abhishek Singh\thanks{Corresponding author. E-Mail: \authorEmail}}
\author[1]{Giovanni Cuozzo}
\author[1]{Alexander Jacob}
\author[1]{Claudia Notarnicola}
\affil[1]{Institute for Earth Observation, Eurac Research, Bolzano, Italy}
\title{Quantum-Inspired Convolutional Networks for High-Resolution Pseudo-Label Generation from Multimodal Earth Observation Data}
\begin{document}
\maketitle

\begin{abstract}
Accurate land‑cover mapping from multimodal Earth observation (EO) data remains a challenge in regions where high‑resolution annotations are scarce or costly to obtain. This work introduces a quantum convolutional encoder–decoder network for pseudo‑label generation using Sentinel‑1 SAR and Sentinel‑2 multispectral imagery, guided only by coarse MODIS‑derived land‑cover maps. The proposed framework uses quantum‑inspired convolutional blocks to enhance feature extraction and cross‑sensor fusion, enabling the model to infer a fine‑scale semantic structure from heterogeneous inputs. Experiments were conducted on the DFC2020 dataset, which provides a quadruple of Sentinel-1 SAR, Sentinel-2 multispectral, low‑resolution MODIS labels and high‑resolution reference land‑cover maps. The proposed training framework first learns a robust representation of multimodal inputs through the reconstruction branch before any supervision from low‑resolution labels is applied. This self‑supervised stage enables the model to capture a fine‑scale spectral–spatial structure, which is then used to generate high‑resolution pseudo‑labels. These pseudo‑labels are subsequently evaluated through classification metrics on the test data, demonstrating that the proposed approach produces more spatially coherent and class‑consistent compared to classical baselines. The results highlight the potential of quantum‑inspired deep learning to improve label generation in low‑annotation remote sensing scenarios, offering a scalable path towards high‑quality land‑cover products for EO applications.
\vspace{-1em}
\end{abstract}
\section{Introduction}
\vspace{-0.5em}
High‑resolution land‑cover mapping plays a key role in environmental monitoring, climate analysis and sustainable land‑use planning, yet its development is often constrained by the limited availability of detailed annotated datasets. Multimodal EO sources such as Sentinel‑1 SAR and Sentinel‑2 multispectral images offer rich and complementary information, but utilizing them effectively requires models capable of learning robust spectral–spatial representations under weak or coarse supervision. Recent advances in deep learning have enabled significant progress in semantic segmentation and data fusion, particularly through encoder–decoder architectures that can reconstruct fine‑scale patterns from heterogeneous inputs.  However, deep convolutional neural networks (CNNs) often require substantial amounts of labeled data, involve large parameter budgets, and may struggle to generalize across varying spectral–spatial distributions, sensor modalities, and acquisition conditions commonly encountered in remote sensing imagery \cite{hua2021semantic, singh2021sigan}. These limitations motivate the exploration of alternative architectures, such as parameter efficient quantum‑inspired convolutional networks that can also enhance feature expressiveness and improve pseudo‑label generation. 
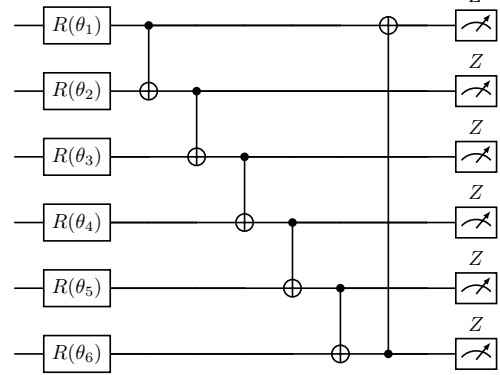
\begin{figure}[h!]
\centering
\vspace{-1.5em}
\resizebox{0.4\textwidth}{!}{
\begin{quantikz}
\lstick{} & \gate{R(\theta_1)} & \ctrl{1} & \qw        & \qw        & \qw        & \qw  & \targ{} &\qw & \meter{Z}\\
\lstick{} & \gate{R(\theta_2)} & \targ{}  & \ctrl{1}   & \qw        & \qw        & \qw  & \qw &\qw & \meter{Z}\\
\lstick{} & \gate{R(\theta_3)} & \qw      & \targ{}    & \ctrl{1}   & \qw        & \qw  & \qw &\qw & \meter{Z}\\
\lstick{} & \gate{R(\theta_4)} & \qw      & \qw        & \targ{}    & \ctrl{1}   & \qw  & \qw &\qw & \meter{Z}\\
\lstick{} & \gate{R(\theta_5)} & \qw      & \qw        & \qw        & \targ{}    & \ctrl{1} & \qw &\qw & \meter{Z}\\
\lstick{} & \gate{R(\theta_6)} & \qw      & \qw        & \qw        & \qw        & \targ{} & \ctrl{-5} & \qw & \meter{Z}
\end{quantikz}
}
\caption{Quantum convolutional circuit (QuantumConv) with $n$ qubits showing staggered entanglement links.}
\vspace{-0.5em}
\label{fig:fig1}
\end{figure}
Building on these limitations, recent studies have highlighted the potential of quantum machine learning (QML) as an emerging alternative for remote sensing applications. QML is a class of machine-learning models based on parametrized quantum circuits (PQCs), which can be trained to perform various downstream tasks such as classification, regression or data generation. Quantum neural networks (QNNs) can be formulated using different architectural styles, including unitary architectures in which the network is composed of parametrized qubit rotations and other unitary operations. In contrast to classical convolutions, which perform linear aggregations of spatially local features, 
a quantum network achieves non-linear feature mixing by encoding information into quantum states through parameterized rotation gates and entangling operations \cite{gong2024quantum,cong2019quantum}. QML models can exploit hybrid quantum–classical training schemes to achieve competitive performance with significantly reduced computational resources, making them an attractive direction for future EO workflows \cite{otgonbaatar2023exploiting}. Moreover, QML offers the advantages of quantum algorithms that can be efficiently simulated on conventional hardware while integrating quantum computational principles with machine learning to address complex optimization and representation‑learning challenges more efficiently than purely classical approaches \cite{rodriguez2025survey, long2025hybrid}. 
In recent years, weak‑supervised and self‑supervised learning strategies have gained prominence as effective alternatives to supervised approaches, particularly in limited, coarser or noisy label EO scenarios \cite{zhou2018brief,zhang2020semisupervised, singh2023refining, 11016942}. These methods exploit coarse labels, unlabeled imagery, or pretext tasks to learn meaningful representations that can generalize across sensors and acquisition conditions. The convergence of parameter efficient quantum‑inspired architectures with weak‑supervised learning provides a promising foundation for developing models capable of generating reliable pseudo‑labels from multimodal satellite data, even when only low‑resolution land‑cover information is available. The main contributions of this paper are as follows;
\begin{enumerate}
    \item We design a quantum-inspired convolutional network, a parameter-efficient quantum convolution–based encoder–decoder for pseudo label generation.
    \item We propose a reconstruction module with an auxiliary classifier that learns spectral–spatial representations and uses the reconstructed features to generate high‑resolution pseudo‑labels.
    \item We compare the effectiveness of the proposed approach with reference to state-of-the-art methods in terms of segmentation accuracies.
\end{enumerate}
The rest of the paper is structured as follows. Section II introduces the proposed approach, quantum convolutional and target use case. Section III highlights the experimental data, loss function formulations, experimental results, and detailed discussions. Finally, Section IV concludes the paper and highlights the future scope.
\vspace{-1em}
\section{Proposed Methodology}
\begin{figure}[t]
    \centering
    \includegraphics[width=.5\textwidth]{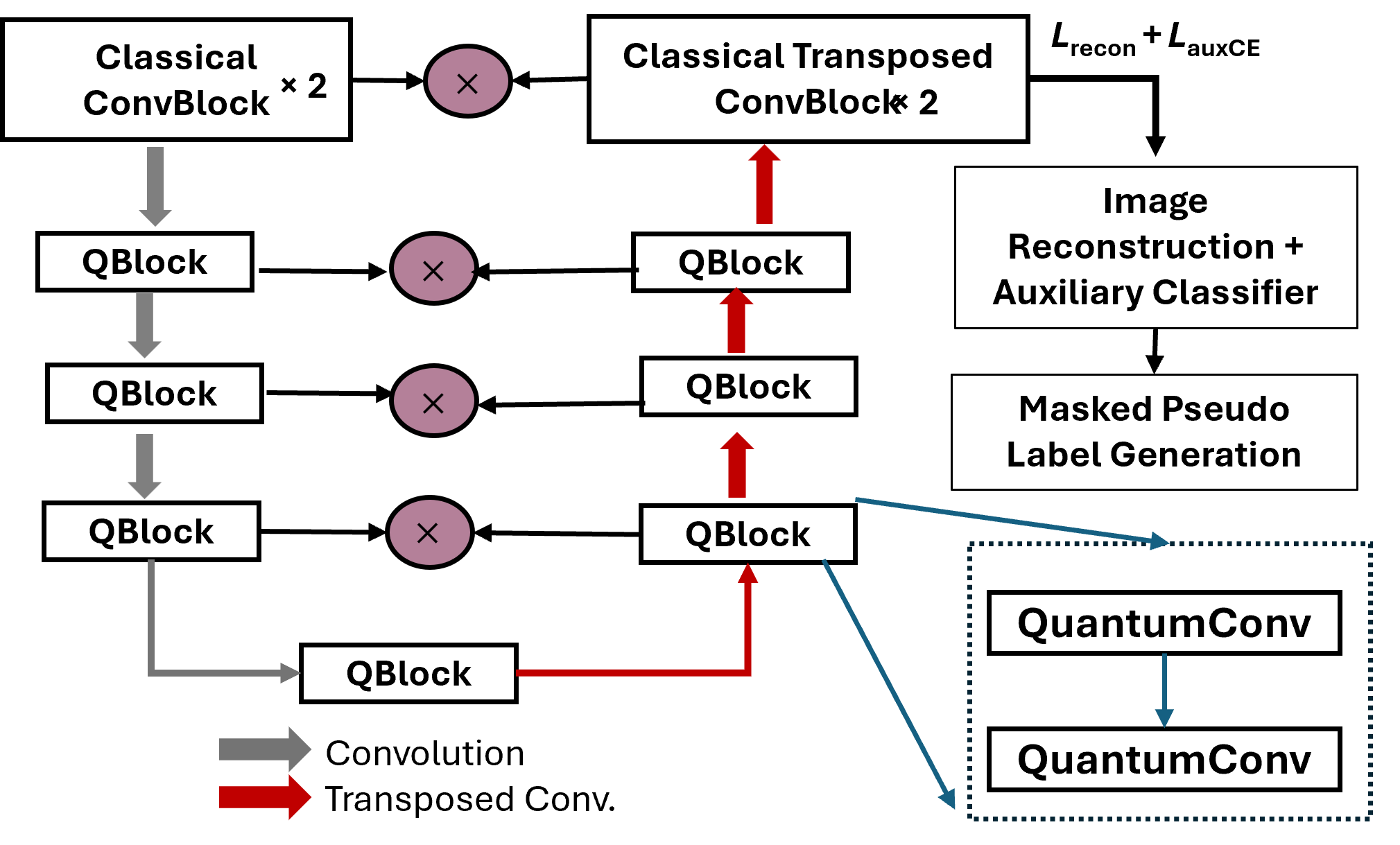}
    \vspace{-2em}
	\caption{Overall framework of the proposed approach for the Land Cover Segmentation with Pre-training for masked pseudo label generation.}
    \vspace{-1em}
	\label{fig:fig2} 
\end{figure}

\vspace{-0.5em}
\cref{fig:fig1} shows the quantum convolutional circuit (QuantumConv) with $6$ qubits showing staggered entanglement links:
$q_1 \!\rightarrow\! q_2$, $q_2 \!\rightarrow\! q_3$, $q_3 \!\rightarrow\! q_4$,  $q_4 \!\rightarrow\! q_5$, $q_5 \!\rightarrow\! q_6$, and $q_6 \!\rightarrow\! q_1$. Each qubit encodes one input feature, undergoes trainable rotations, and participates in global entanglement prior to measurement. \cref{fig:fig2} further details the overall framework and integration of QuantumConv in Quantum Blocks (QBlock) in the proposed network for pseudo-label generation.
The proposed QuantumConv layer uses a parameterized quantum circuit (PQC) with a depth of 4 layers ($L=4$) and a ring entanglement pattern implemented by $StronglyEntanglingLayers$.
For each spatial location, a non-overlapping $2 \times 2$ patch is extracted from the intermediate classical feature map. Rather than directly encoding all spatial elements, the patch is first aggregated by computing the channel-wise mean over the $2 \times 2$ region. This operation produces a compact local descriptor $\mathbf{x} \in \mathbb{R}^{C}$, where $C$ denotes the number of input channels. To interface with the quantum circuit, this vector is adjusted to match the number of qubits optimised to $n=6$. If $C < 6$, zero-padding is applied; if $C > 6$, the vector is truncated to its first six components. This results in an input vector $\mathbf{x} \in \mathbb{R}^{6}$, where each element is mapped to the rotation angle of a corresponding qubit by angle encoding. A 6-qubit PQC is then defined, where qubit $q_i$ encodes feature $x_i$. These quantum transformations generate non-linear feature interactions in Hilbert space \cite{young1988introduction}, enabling enhanced representation learning across channels.
The overall circuit transformation is given by:
\begin{equation}
|\psi_{\text{out}}\rangle =
U_{\text{total}}(\Theta)\,
U_{\text{embed}}(\mathbf{x})\,
|0\rangle^{\otimes 6},
\end{equation}
where $U_{\text{embed}}(\mathbf{x}) = \bigotimes_{i=1}^{6} R_Y(x_i)$ denotes the classical-to-quantum embedding through single-qubit $R_Y$ rotations, $U_{\text{total}}(\Theta)$ represents a stack of $L$ parametrized entangling layers, and $|0\rangle^{\otimes 6}$ is the ground state initialization. For layer $l$, the parametrized transformation followed by entanglement is expressed as:
\begin{equation}
U^{(l)}(\Theta^{(l)}) =
U_{\text{ent}}
\left(
\bigotimes_{i=1}^{6}
R_Z(\theta^{(l)}_{i,1})
R_Y(\theta^{(l)}_{i,2})
R_Z(\theta^{(l)}_{i,3})
\right),
\end{equation}
where $R_Z(\cdot)$ and $R_Y(\cdot)$ are single-qubit rotation matrices and $U_{\text{ent}}$ denotes an entangling operator implemented using a ring topology of CNOT gates. The circuit concludes with measurements in the Pauli-$Z$ basis:
\begin{equation}
z_i = \langle \psi_{\text{out}} | Z_i | \psi_{\text{out}} \rangle,
\qquad
Z =
\begin{bmatrix}
1 & 0 \\
0 & -1
\end{bmatrix}.
\end{equation}
The resulting vector $\mathbf{z} = [z_1, z_2, \ldots, z_6]^{\mathsf{T}}$ represents nonlinear quantum features arising from entanglement and quantum interference effects. Given an input image $I \in \mathbb{R}^{P \times Q \times R}$, the overall model follows a hybrid quantum–classical encoder–decoder architecture. Classical convolutional layers first extract structured low-level spatial representations. 
The QuantumConv layers are then applied within seven quantum-enhanced feature extraction blocks (QBlocks) distributed across the encoder and decoder stages. Specifically, the architecture incorporates three QBlocks in the encoder to progressively refine low- to mid-level representations, one QBlock at the bottleneck to enhance global contextual feature learning, and three additional QBlocks in the decoder to support accurate feature reconstruction and class-specific discrimination. This symmetric distribution was empirically optimized to balance representational capacity and computational overhead, enabling improved hierarchical feature extraction while maintaining stable training dynamics. The inclusion of quantum-enhanced processing at multiple resolution levels allows the network to capture complex correlations through entanglement-driven non-linear transformations, thereby strengthening both encoding and decoding networks.

Each QBlock contains two QuantumConv layers, where local spatial patches are aggregated, encoded into a 6-qubit PQC, and transformed through parametrized entangling operations. 
The quantum outputs correspond to expectation values of Pauli-Z measurements which are projected into the classical feature space through a learnable linear layer and normalization, enabling integration with the subsequent convolutional operations. After the final QBlock, the architecture returns entirely to classical processing, applying a conventional convolutional layer before branching into task-specific output heads. To efficiently exploit low-resolution annotated data, the proposed approach is trained in two sequential phases: image reconstruction and auxiliary learning for masked pseudo-label generation and semantic segmentation. The proposed strategy allows the network to first learn robust feature representations in a self-supervised manner, then refine pseudo labels for weak supervision, and finally adapt to dense segmentation with limited reliance on high-quality annotations. The first phase aims to exploit both low-level image features and semantic consistency to generate reliable pseudo labels. The proposed approach is trained in a multitask setting that jointly reconstructs the input imagery and predicts the probabilities of the auxiliary classes. The reconstruction branch encourages the encoder to preserve fine-grained spatial and spectral structures, improving feature generalization, and reducing domain gaps between low- and high-resolution representations. Concurrently, the auxiliary classification branch learns semantic information from coarse labels, producing temperature-scaled confidence maps that capture category-level consistency.
\vspace{-1.5em}
\section{Experimental Results and Discussions}
\vspace{-1em}
\subsection{Experimental Data and Model Setup and Loss Function Formulation}
\vspace{-0.5em}
We have considered open source DFC2020 \cite{rha7-m332-19} datasets. The experiments were conducted on the open-source DFC2020 dataset  in which we have considered low-resolution 500m-MODIS-derived land-cover maps and semi-manually derived high-resolution land-cover maps. This dataset originally consisted of a quadruple of Sentinel-1 SAR data, Sentinel-2 multispectral imagery, low-resolution land cover and corresponding high-resolution land cover maps. Each patch size is $256 \times256$ pixels. We have used $4615$, $513$ and $986$ patches from the DFC2020 dataset to train, validate and test the proposed approach. To align the heterogeneous label sources within the DFC2020 dataset, we first remapped the original 17 class MODIS‑derived LR land‑cover labels into a unified 10 class scheme that matches the semantic structure of the high‑resolution DFC2020 annotations. The model is trained using the selected 5 dominant classes with remapped labels like Forest, Grassland, Croplands, Urban and Water and evaluated on these classes on the test set.
\begin{figure}[t]
    \centering    \includegraphics[width=.5\textwidth]{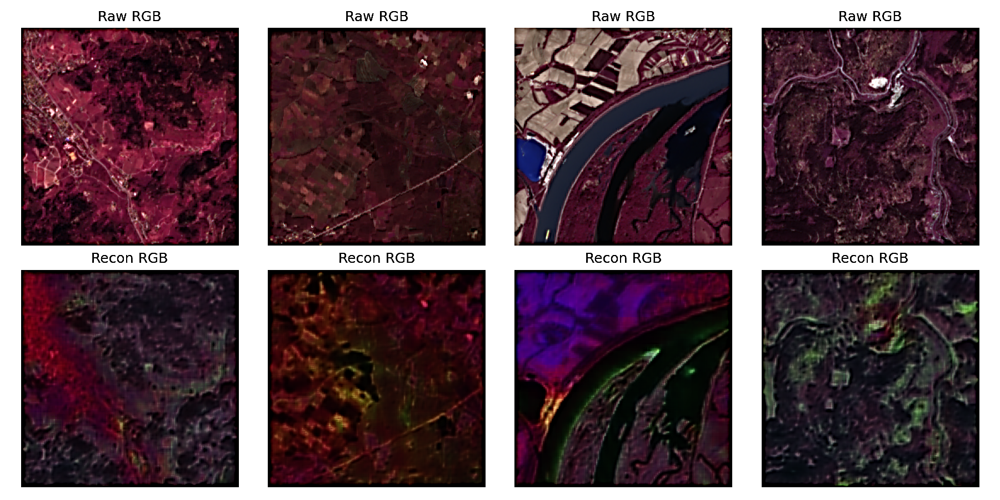}
	\caption{Pre-training for Image Reconstruction and Raw vs Reconstructed patch when both Reconstruction and Auxiliary Classifier branch were trained}
	\label{fig:fig3} 
    \vspace{-1em}
\end{figure}
For model training, Adam (Adaptive Moment Estimation) \cite{adam2014method} was used with a learning rate of $0.003$.
During experimental training, the batch size is set to $64$, and all models are trained. The Adam update rule for a parameter vector $\theta$ with learning rate $\eta$, decay rates $\beta_1$ and $\beta_2$, and small constant $\epsilon$ can be expressed as $\theta_{t+1}= \theta_t - \eta \frac{\hat{m}_t}{\sqrt{\hat{v}_t} + \epsilon}$, where $\hat{m}_t = \frac{m_t}{1 - \beta_1^t}$ and $\hat{v}_t = \frac{v_t}{1 - \beta_2^t}$ are bias-corrected estimates of the moving average of the gradients, $m_t = \beta_1 m_{t-1} + (1 - \beta_1) g_t$ and $v_t = \beta_2 v_{t-1} + (1 - \beta_2) g_t^2$, $g_t=\nabla_{\theta} J(\theta_t)$ is the gradient of the cost function $J(\theta)$ with respect to the parameter vector $\theta$ at the current time step $t$.

The reconstruction head aims to restore the original multimodal input, encouraging the encoder to retain fine spatial and spectral structures. To achieve this, we employ a hybrid L1-SSIM loss and the overall reconstruction loss, $\mathcal{L}_{recon} = \alpha \, \| \hat{X}_{rec} - X_{raw} \|_1 + (1 - \alpha) \, [1 - \text{SSIM}(\hat{X}_{rec}, X_{raw})]$, where $\text{SSIM}(\cdot)$ denotes the Structural Similarity Index and $\alpha$ balances pixel-level and perceptual fidelity. The auxiliary branch promotes the alignment of semantic features and confidence calibration in the early stages of training. 
\begin{figure}[t]
    \centering    \includegraphics[width=.5\textwidth]{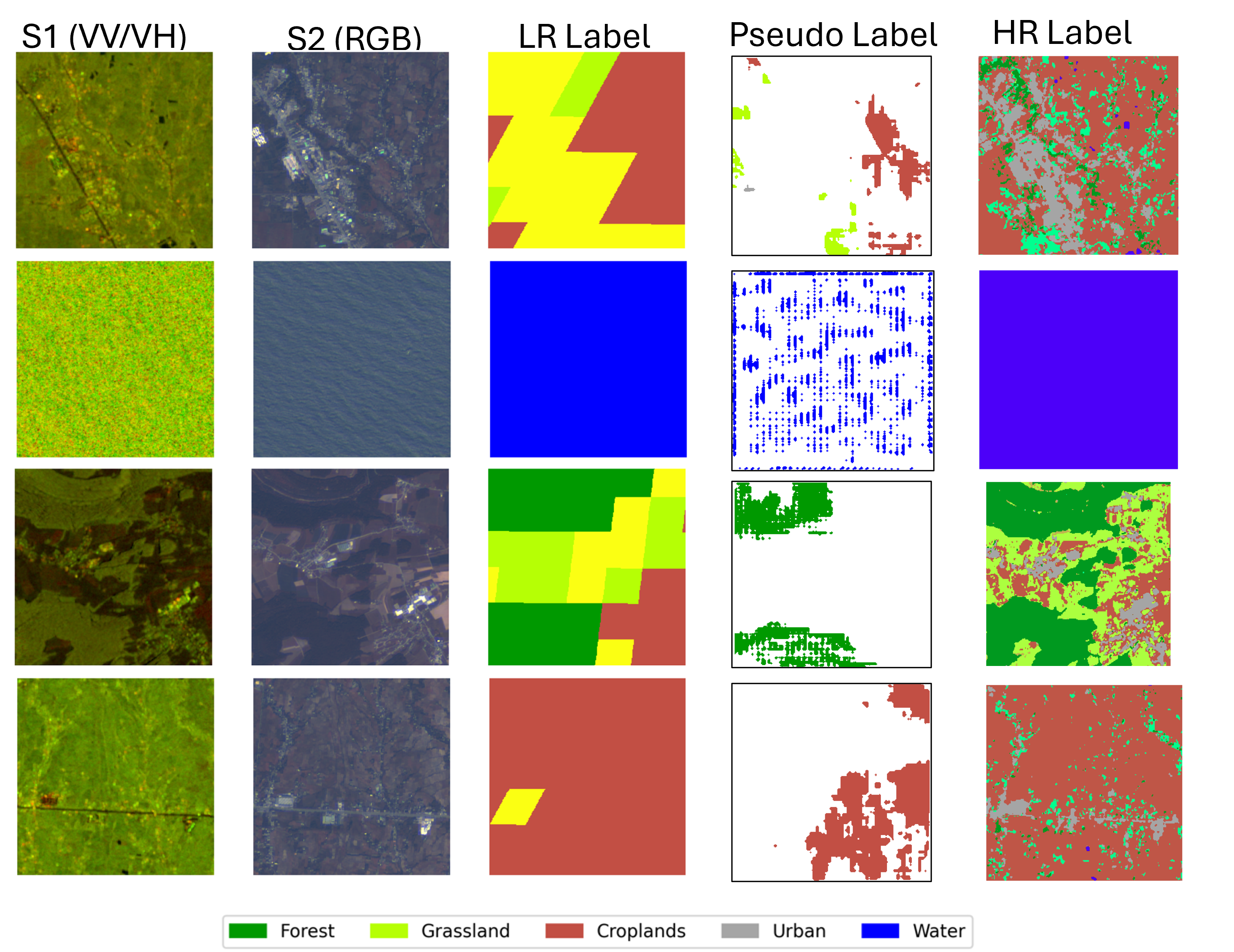}
	\caption{Pre-training for High-Resolution Pseudo Label Generation}
	\label{fig:fig4} 
    \vspace{-1em}
\end{figure}
\begin{table}[]
\centering
\caption{Ablation study of the proposed approach on the selected 5 dominant classes of the DFC2020 test set with remapped labels
like Forest, Grassland, Croplands, Urban and Water. Experiments used a batch size of 64 over 986 image patches of size $256 \times256$. Baseline - Classical Convolutional Blocks, Qbottleneck - Only Quantum-inspired bottleneck }
\label{tab:tab1}
\resizebox{0.5\textwidth}{!}{%
\begin{tabular}{cccc}
\toprule
\textbf{Metrics (\%)} & \textbf{Baseline} & \textbf{Qbottleneck}  & \textbf{Proposed}\\
\midrule
\textbf{Overall Accuracy (OA)}      &0.7993 &0.8171  & \textbf{0.8314}\\ 
\textbf{Kappa Coefficient ($\kappa$)}     &0.7163 &0.7413    & \textbf{0.7612}\\ 
\textbf{$mIoU$}                &0.5998   &0.6238 &\textbf{0.6434}\\
\hline
 \textbf{Total No. of Parameters (approx.)}&2,007,898  & 1,126,378 & 169,546\\ 
\hline
\vspace{-2em}
\end{tabular}%
}
\end{table}
The auxiliary loss, $\mathcal{L}_{CEaux} = - \sum_{h=1}^{C} \gamma_h \, \log(\hat{\gamma}_h)$, where $C$ is the number of classes, $\gamma_h$ is the true probability of class $h$, and $\hat{\gamma}_h$ is the predicted probability for class $h$.  
Label smoothing is applied with the factor $\epsilon$ to prevent overconfident predictions and improve pseudo-label calibration. During initial epochs, both branches are optimized jointly using a weighted total loss, $\mathcal{L}_{total} = w_{recon} \, \mathcal{L}_{rec} + w_{aux} \, \mathcal{L}_{aux}$, where $w_{rec}$ and $w_{aux}$ are the weighting factors that balance the reconstruction and the auxiliary classification. For the downstream segmentation task, the network is optimized using a hybrid segmentation loss that combines label-smoothed cross-entropy and Dice loss. The cross-entropy term promotes pixel-wise classification accuracy while the Dice term improves overlap-based region consistency and mitigates class imbalance. The gradients of the PQC are computed through automatic differentiation using PennyLane's differentiable QNode ($interface="torch", diff-method="backprop"$), which is fully integrated with the PyTorch autograd framework. To experimentally verify gradient propagation, we monitored the gradients of all trainable PQC parameters during training. The results demonstrated non-zero gradients throughout the quantum layers, confirming successful end-to-end gradient flow and optimization of the hybrid quantum-classical architecture. While gradient magnitudes progressively decrease in deeper quantum blocks, indicating a vanishing-gradient tendency, the PQC parameters remain trainable and are updated through standard backpropagation. The experiments were conducted on NVIDIA $A100D-4-40C$ GPU under CUDA Version: $12.4$, Ubuntu 22.04.5 LTS (GNU/Linux 6.8.0-87-generic x86-64). We have used pennylane $0.38.0$, torch $2.6.0+cu124$, rasterio $1.4.3$, matplotlib $3.7.5$ and numpy $1.26.4$ to implement the proposed approach and reference methods.
\subsection{Qualitative and Quantitative Evaluation and Discussions}
\cref{fig:fig3} shows the pre-training stage for the generation of masked pseudo-labels where the reconstruction and auxiliary branches of the proposed approach were trained simultaneously. During this stage, the network receives high-resolution input patches and learns to reconstruct them through a quantum-enhanced encoder–decoder architecture, ensuring that the latent representations preserve fine-grained spatial and spectral details. In parallel, the auxiliary branch provides additional supervision that regularizes the latent space by enforcing semantic consistency across the reconstructed and masked regions. This dual-branch learning allows the proposed approach to develop robust feature embeddings even without accurate ground-truth labels. \cref{fig:fig4} highlights the generation of masked labels from the input high resolution image. As shown in \cref{fig:fig4}, the self-supervised trained reconstruction branch is subsequently used to produce masked pseudo labels by selectively masking regions in the input image and then reconstructing them through the learned latent representations. This process ensures that the model benefits from rich contextual information derived directly from the imagery, hence improving label quality and their efficacy tested by segmentation performance. \cref{tab:tab1} highlights the ablation study of the proposed approach on the DFC2020 test set.
\begin{table}[]
\centering
\caption{Comparison of classification results between the proposed method and reference approaches on the selected 5 dominant classes of the DFC2020 test set with remapped labels like Forest, Grassland, Croplands, Urban and Water. Experiments used a batch size of 64 over 986 image patches of size $256 \times256$.}
\label{tab:tab2}
\resizebox{0.5\textwidth}{!}{%
\begin{tabular}{cccc}
\toprule
\textbf{Metrics (\%)} & \textbf{UNet} & \textbf{TransUNet}  & \textbf{Proposed}\\
\midrule
\textbf{Overall Accuracy (OA)}      &0.8194 &0.8238  & \textbf{0.8314}\\ 
\textbf{Kappa Coefficient ($\kappa$)}     &0.7454 &0.7500    & \textbf{0.7612}\\ 
\textbf{$mIoU$}                &0.6272      &0.6154 &\textbf{0.6434}\\
\hline
 \textbf{Total No. of Parameters (approx.)}& 31,392,395 & 26,665,867 &169,546\\ 
\hline
\textbf{Computational Complexity}         & $1.64$         & $1.10$    & \textbf{$2.12$} \\
\textbf{(Training: iterations/s)}              &       &      &     \\ 
\hline \hline
Classes & & Class-wise F1 Score     & \\
\hline
Forest       &\textbf{0.7689}     &0.5908    & 0.7163    \\
Grassland        &0.5900     &0.6645  & \textbf{0.6798}     \\
Crop Lands           &0.6947  &0.6969  & 0.6940     \\
Artificial Lands           &0.7575  &0.7891  & \textbf{0.7931}     \\
Water           &0.9748  &\textbf{0.9868}  & 0.9771     \\
\hline
\end{tabular}%
}
\vspace{-1em}
\end{table}
\cref{tab:tab2} highlights the comparison of the classification results of the proposed approach with the reference methods like UNet and TransUNet on the test set of the DFC2020 data. UNet used convolutional encoder-decoder architecture with skip connections for dense prediction, whereas TransUNet combines a CNN encoder with Transformer-based global feature modeling. All methods were trained from scratch using the same training and test partitions, input patch size, batch size and loss function. The baseline implementations were configured following their standard architectures and adapted only to match the input dimensionality and number of output classes of the dataset. The results demonstrate that the proposed approach consistently outperforms the baseline approaches across most evaluation metrics, confirming the effectiveness of the quantum-enhanced feature extraction and the masked pseudo-label pre-training strategy. In particular, the pseudo-labels generated during the pre-training stage were evaluated not only through overall segmentation accuracy. The experimental results demonstrate that the proposed approach architecture effectively integrates quantum feature extraction with classical convolutional learning, resulting in improved segmentation performance across several land-cover classes. The use of masked pseudo-label generation through self-supervised reconstruction to more stable representation learning, particularly in regions with low-resolution annotations. The auxiliary branch further reinforces the ability of the proposed approach to disentangle class-specific patterns, thereby enhancing the robustness and consistency of predictions in different spatial conditions. The proposed quantum components are intended to investigate the potential of hybrid quantum-classical networks rather than to claim universal superiority over lightweight classical architectures. The PQC layers introduce a distinct feature transformation mechanism based on parameterized quantum operations and entanglement while maintaining a relatively small number of trainable parameters. Nevertheless, efficient classical alternatives remain strong competitors, and a comprehensive comparison with lightweight CNN modules under comparable parameter and computational budgets is an important direction for future work. In terms of computational complexity shown in \cref{tab:tab2}, the proposed method achieves the best overall segmentation performance while using a substantially more compact model than the compared architectures. Although the hybrid quantum-classical design introduces additional computational overhead during training due to quantum circuit simulation, it offers a favorable trade-off between model compactness and predictive performance. Therefore, the primary practical advantage of the proposed approach lies in its parameter efficiency and accuracy rather than raw computational speed.

\section{Conclusion}
\vspace{-0.5em}
This paper presented a quantum‑integrated convolutional encoder–decoder framework for generating high‑resolution pseudo‑labels from multimodal Sentinel‑1 and Sentinel‑2 images using only coarse MODIS land‑cover information. Experiments on the DFC2020 dataset demonstrated that the proposed model achieves competitive and often outperforms compared to the reference methods, despite using fewer parameters. The model shows a significant improvement in accuracy metrics, confirming its effectiveness for land‑cover mapping. Due to its lightweight and parameter‑efficient design, the proposed architecture shows strong potential for deployment as a compact onboard AI model for real‑time EO applications.
\begin{acknowledgments}
\vspace{-0.5em}
The authors acknowledge the scientific context and discussions with Francesco Rossi associated with the ESA PODIUMS project (Payload for Onboard Data and Intelligent Multispectral Systems).
\end{acknowledgments}
\vspace{-1em}
\printbibliography
\addcontentsline{toc}{section}{References}

\end{document}